\documentclass[preprint,12pt,]{elsarticle}

\usepackage{setspace}
\biboptions{sort&compress}
\usepackage[numbers, sort&compress]{natbib}
\usepackage{amsmath,amssymb,amsfonts,mathtools,bm}
\usepackage{algorithm,algpseudocode}
\usepackage{graphicx}
\usepackage{array}
\usepackage{stfloats}
\usepackage{url}
\usepackage{verbatim}
\usepackage{graphicx}
\usepackage{booktabs,multirow}
\usepackage{tabularx}
\newcolumntype{Y}{>{\centering\arraybackslash}X}
\usepackage{threeparttable}
\usepackage{textcomp}
\usepackage{xcolor}
\usepackage{enumerate}
\graphicspath{{pic/}}
\usepackage{subcaption}
\usepackage{bbding}

\journal{Sustainable Energy Grids \& Networks}

\begin{document}

\begin{frontmatter}



\title{Noise-Aware Bayesian Optimization Approach for Capacity Planning of the Distributed Energy Resources in an Active Distribution Network}


\author[label1]{Ruizhe Yang}\author[label1]{Zhongkai Yi}\author[label1]{Ying Xu}\author[label1]{ Dazhi Yang}\author[label1]{Zhenghong Tu}

\affiliation[label1]{
            organization={School of Electrical Engineering and Automation},
            addressline={92 Xida St, Nangang}, 
            city={Harbin},
            postcode={150001}, 
            state={Heilongjiang},
            country={China}}

\begin{abstract}
The growing penetration of renewable energy sources (RESs) in active distribution networks (ADNs) leads to complex and uncertain operation scenarios, resulting in significant deviations and risks for the ADN operation.
In this study, a collaborative capacity planning of the distributed energy resources in an ADN is proposed to enhance the RES accommodation capability. The variability of RESs, characteristics of adjustable demand response resources, ADN bi-directional power flow, and security operation limitations are considered in the proposed model.
To address the noise term caused by the inevitable deviation between the operation simulation and real-world environments, an improved noise-aware Bayesian optimization algorithm with the probabilistic surrogate model is proposed to overcome the interference from the environmental noise and sample-efficiently optimize the capacity planning model under noisy circumstances. 
Numerical simulation results verify the superiority of the proposed approach in coping with environmental noise and achieving lower annual cost and higher computation efficiency.
\end{abstract}




\end{frontmatter}


\section{Introduction}\label{S1}
Active distribution networks (ADNs), as opposed to the traditional passive ones with a power-receiving nature, play a vital role in transforming the current power system into a low-carbon one \cite{9869336}. In a nutshell, ADNs are distribution networks with distributed energy resources (DERs), which permit energy exchange through various control mechanisms and a flexible network topology. Amongst numerous development needs, the allocation of DERs' capabilities ought to be regarded as the foremost one, which is what concerns this work. 

The growing participation of renewable energy sources (RESs) in ADNs raises safety and stability concerns, due to the inevitable intermittency and variability in the output power of RESs \cite{9696337}. To tame such intermittency and variability and to improve the resilience and security of ADNs, the pairing of RES with energy storage systems (ESSs) has been advocated in both distribution and transmission networks, despite the high cost of ESSs, which has hitherto been limiting the installed capacity of ESSs \cite{rahman2020assessment}. As such, the balance between the energy economy and safety has to be factored in during the planning of ADNs. Another major challenge associated with having RESs configured in an ADN is the sheer number of scenarios involved. Since the scenarios are too numerous for a capacity planning model to handle, it is common to observe deviation between the real operations from the simulated ideals, which confront most of the current capacity planning approaches. Although several proposals have claimed to offer solutions to the problem, most still face the dilemma of sacrificing the computation efficiency or the feasibility of the final capacity allocation scheme \cite{9990600,deng2020power}. As a result, a novel method of modeling the simulation deviation caused by the RES variability is urgently required. In light of the requirement mentioned above, this work proposes a modeling method for the simulation deviation and a solution method based on Bayesian optimization (BO) to juggle the computation efficiency and the feasibility of the final capacity allocation scheme. 

\subsection{Literature Review and Research Gaps}

Capacity planning in a power system setting refers to the allocation of resources for serving the load demand. In general, the capacity planning of power systems can be divided into two classes, one being macro-level planning on yearly or daily scales and the other being integrated planning considering the RES variability at hourly or even smaller time scales. For macro-level configurations, as typified by \cite{wierzbowski2016milp,guo2016multi,9769874,dhakouani2017long}, the primary attention is paid to the long-lead-time factors such as the morphing of load demand, the transition of energy mix, or the policies concerning energy subsidies and national targets. In contrast, integrated planning studies place more emphasis on detailed simulation of the operation with some specific characteristics or elements considered \cite{xuan2022two,liang2021data,9801660,9760012}. For instance, the involvement of carbon capture power plants in an electricity-gas system is stressed in \cite{xuan2022two}; the wind farm planning method is proposed in \cite{9801660} considering the spatial smoothing effect; and considering the popularity of electric vehicles, a planning strategy for fast-charging stations is put forward in \cite{9760012}.

Compared to the above-mentioned planning models, the capacity planning of an ADN involves more elements due to the integration of various DERs, electronic devices, and demand response mechanisms. Various models have been proposed to address the salient characteristics of ADNs. For instance, a planning method for an ADN with a hybrid topology was proposed by \citet{huang2021bi} to determine the optimal location and sizing of converters and select line types and states. A formulation for the reconfiguration problem that accounts for partial restoration scenarios where the whole unsupplied area cannot be restored has been presented in \cite{sekhavatmanesh2020multi}. \citet{nick2017optimal} designed a procedure for the optimal siting and sizing of ESSs owned and directly controlled by ADN operators. In \cite{al2013planning}, an approach to planning the sizing of DER units was proposed, considering the influence of multi-DER configurations. 

In the literature concerning the capacity planning of RESs, the methods for dealing with the simulation deviation caused by RESs' variability have been discussed considerable times \cite{9790344,wang2022novel,chicco2012electrical,zhang2022efficient,javed2021economic,bhattacharjee2020benefits}. These previous works on eliminating or diminishing the deviation can be classified into two categories, one relies on the \textit{typical scenarios} to conclude all the variability into a certain set of RES scenarios \cite{9790344,wang2022novel,chicco2012electrical}, which is easy to be carried out and contributes to the decreasing of the planning model's complexity. However, there is no consensus on the standard and method of selecting the typical scenarios, indicating that the objectivity and feasibility of such a methodology are questionable. The other counts on \textit{numerous scenarios} \cite{zhang2022efficient,javed2021economic,bhattacharjee2020benefits}, which manages to incorporate as many RES scenarios as possible by prolonging the dispatch cycle to one year or even longer. The augmentation of scenario numbers successfully reduces the simulation deviation but also requests simplifications on the capacity planning model to make it less computationally intensive.

As for the solution methods, the DER capacity planning problems can be solved by analytical methods, such as stochastic programming\cite{9908163}, robust optimization\cite{xuan2022two}, or other kinds of mathematical programming algorithms. In addition, heuristic methods \cite{9920235,javed2021economic,8891057} also can be adopted if there exist nonconvex or nonlinear elements in the mathematical problem. Nevertheless, considering the uncertainty of renewable energy, time-varying demand-side resources, and inaccurate network parameters, an ADN is actually operated in a complex environment with randomness, volatility, and imprecision. Traditional mathematical methods become ineffective in the mathematical modeling and decision-making processes, since it is difficult to formulate an accurate mathematical model of the ADN environment. As such, heuristic methods such as particle swarm optimization (PSO) and history-driven differential evolution \cite{9920235} are adopted to solve the ADN capacity planning problem. These methods alleviate the problem of inaccurate modeling to some extent, whereas heuristic methods themselves suffer from low optimizing efficiency, incapacity in handling environmental noise, and premature convergence.

Although some preliminary investigations on ADN capacity planning concerning RESs have been conducted. There are still several unavoidable drawbacks to the current approaches.
Firstly, ADN capacity planning should face the uncertain future RES scenario. The deviation between the ADN simulation and the real situation is bound to hold. Such deviation can hardly be modeled analytically, resulting in the futility of mathematical programming algorithms.
As for the heuristic approaches, they are also limited in handling the deviation and are likely to be inefficient, especially when multiple decision variables exist.
Secondly, the variability of RES leads to infinite possible scenarios, resulting in deviations of the ADN simulation, as it is impossible to consider all scenarios in one model. Certain simplifications in planning models or RES scenarios themselves have to be done to avoid the immense computation complexities engendered by the huge number of scenarios.
Last but not least, few available approaches concentrate on the direct modeling of the deviation of operation simulation, and the research on the ADN network topology and power flow security constraints is worth further in-depth analysis as well.

\subsection{Summary of Major Contributions}

To address the above issues, an operation simulation based capacity planning approach is proposed along with a noise-aware Bayesian optimization (NBO) algorithm, which is utilized to solve the optimization of the annual cost of an ADN by determining the optimal capacities of the DERs contained in it, including distributed RESs and ESSs.

Unlike traditional capacity planning models, the proposed approach does not attempt to eliminate or diminish the deviation from the ADN simulation caused by RES variability, but instead formulates the deviation as a noise function added to the objective function of the planning. All the outcomes from the ADN simulation model turn into noisy accordingly, and the adopted optimization algorithm, NBO, has been certified to be perfectly adaptive to such noisy circumstances with the capability to accomplish the optimization efficiently and reliably.

Compared with existing works, three main contributions of the article are summarized as follows:

\begin{itemize}
\item[i)] By incorporating the probabilistic inference model of the optimization objective and noisy expected improvement function, an \textit{improved} noise-aware Bayesian optimization algorithm is proposed to overcome the interference from the environmental noise. The proposed approach effectively addresses the noise term caused by the inevitable deviation between the simulation and real-world environments. Compared with the traditional Bayesian optimization and heuristic algorithms \cite{javed2021economic,8891057}, the proposed approach achieves enhanced optimality and higher computation efficiency.

\item[ii)] The proposed NBO algorithm is employed in the DER capacity planning problems in ADNs \textit{for the first time}. Compared with the existing scenario-based capacity planning approaches \cite{nick2017optimal,sekhavatmanesh2020multi}, the proposed approach achieves an economical and environmentally friendly planning result without relying on the repeated simulation of numerous scenarios.

\item[iii)] The correlation between RESs and ESSs along with the characteristics of ADNs are considered in the capacity planning model, including the bi-directional power flow constraints, voltage stability constraints, and incorporation of the dispatchable demand response resources. The approach efficiently guarantees an ADN's stable and economic operation and enhances the RES accommodation through the distributed capacity planning of DERs compared to the traditional centralized planning approaches.
\end{itemize}

The rest of the article is organized as follows. The prerequisite knowledge and framework of the proposed approach are introduced in Section \ref{S2}. The formulation of the capacity planning model is presented in Section \ref{S3}. The mathematical details of the solution method NBO are presented in Section \ref{S4}. Case studies and conclusions are provided in Section \ref{S5} and Section \ref{S6}, respectively.

\section{Problem Formation}\label{S2}

Considering the diversity of ADNs, the specific structure of ADNs that the study concentrates on is introduced in the section, along with the analysis of simulation deviation and framework of the proposed capacity planning approach.

\subsection{Typical ADN Structure}
\begin{figure}[!htbp]
	\centering
	\includegraphics[scale=1]{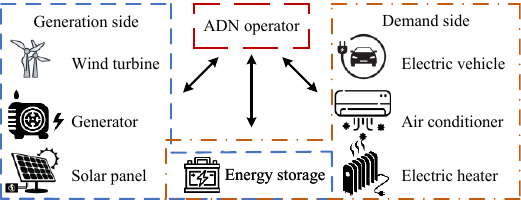}
	\caption{The structure of a typical type of ADN.}
	\label{ADN}
\end{figure}

With all the multifarious elements that an ADN might contain, the study concentrates on one of the most common ADN types, which is composed of wind turbines, solar panels, diesel generators, ESSs, and demand response resources like electric vehicles, air conditioners, and so on. The ADN elements are depicted in Fig. \ref{ADN}. The proposed capacity planning approach is dedicated to seeking the optimal combination of elements belonging to the generation side, excluding diesel generators.
\subsection{Simulation Deviation of ADN Operation}

In the assessment of various allocation schemes, it is imperative to conduct simulations of the ADN operations under different proposed schemes. Nonetheless, since all DERs waited for planning are expected to be operational in the future, the inherent unpredictability and boundlessness of long-term future RES scenarios introduce an unavoidable bias in the evaluation process. While historical data undeniably harbors valuable insights into future trends, prevailing research methodologies tend to employ this data as a proxy for future conditions, often overlooking the deviation that this approach may introduce. In contrast to this trend, this study confronts the deviation head-on. The deviation is modeled into a noise function and its side effects are properly handled by the proposed approach, ensuring that the capacity scheme derived from the proposed planning model exhibits enhanced performance in the actual operational environment.

\subsection{Framework of the Proposed Noise-Aware Bayesian Capacity Planning Approach}
\begin{figure*}[htb]
	\centering
	\includegraphics[scale=0.85]{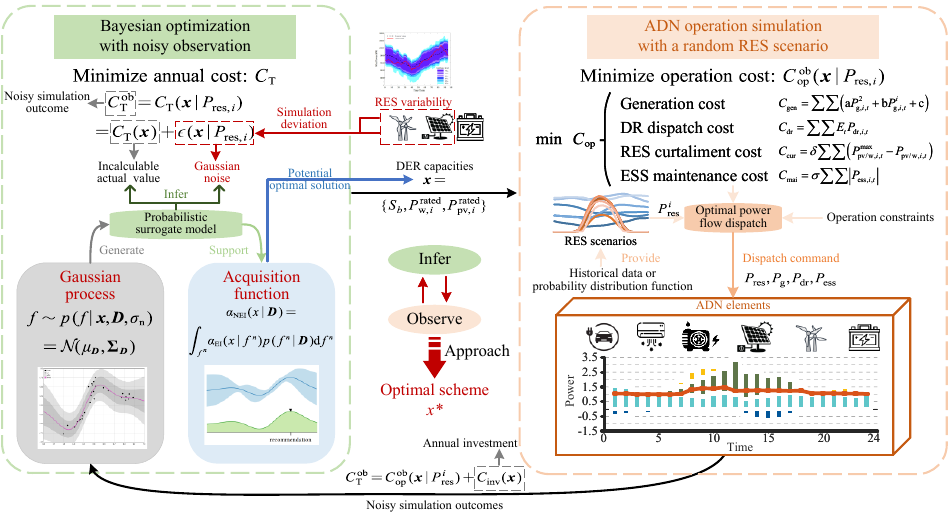}
	\caption{The framework of the proposed NBO based DERs collaborative capacity planning approach}
	\label{framework}
\end{figure*}
The proposed noise-aware Bayesian based capacity planning approach, depicted in Fig. \ref{framework}, consists of an optimization model based on the NBO algorithm and an operation simulation model. 
The noise-aware Bayesian based capacity planning approach works similarly to the traditional bi-level capacity planning model. At first, the ADN operation simulation model receives the capacities of each DER to be planned as a capacity allocation scheme from the optimization model. Then, it generates the optimal dispatch commands for the elements within an ADN and evaluates its annual operation cost. Finally, the cost is reintegrated into the optimization model, augmented by the annual investment, to yield the aggregate annual cost.
With all the capacity allocation schemes and their corresponding cost from each iteration, deemed as the outcomes from the black box objective function, the Gaussian process (GP) model is used to infer the underlying relationship between the DERs' capacities (decision variables) and the annual cost (objective function) by generating a probabilistic surrogate model of the objection function. 
Most importantly, the inference from surrogate model can keep itself updated and improve its accuracy after each iteration with the help of extra outcomes. 
Guided by the surrogate model, the search of the optimal solution is conducted by ensuring that the candidate scheme to be evaluated in the simulation model possesses the maximal mathematical expectation in the improvement of optimal value under the current probabilistic surrogate model. The exact process is achieved by maximizing an acquisition function.
This approach is proven to have a faster convergence rate and is more adaptive to noisy conditions. After sufficient iteration, the NBO algorithm can approximate the optimal capacity allocation scheme, ensuring the lowest annual cost of an ADN.

\section{DER Collaborative Capacity Planning Model in an ADN}\label{S3}

In this section, a detailed explanation of the two main components of the DER collaborative capacity planning model is provided. The first subsection outlines the final objective of the capacity planning model, whereas the second subsection proposes the ADN operation simulation model with the safety constraints on voltage stability and coordination of the DERs and demand response resources to imitate the actual operation situation. At last, the modeling of the ADN simulation deviation and establishment of the noise function to describe the deviation are introduced.

\subsection{Optimization Model}\label{AA}

The optimization model takes the annual cost of an ADN as the objective function to minimize, which can be divided into the equivalent annual investment cost of DERs ($C_{\mathrm{inv}}$), the generation cost of diesel generators ($ C_{\mathrm{gen}}$), the compensation for the dispatch of demand response resources ($C_{\mathrm{dr}}$), the punishment of RES power curtailment ($C_\mathrm{cur}$), and the maintenance cost of ESSs ($C_\mathrm{mai}$) in an ADN. It should be noted that, except for $C_{\mathrm{inv}}$, all other costs are related to the operation of an ADN, and thus are covered in the next subsection. If the operation cost is denoted as $C_{\mathrm{op}}$, the total annual cost of an ADN is:
\begin{equation}\label{obj}
	\min \phantom{1} C_{\mathrm{T}}=C_{\mathrm{op}}+C_{\mathrm{inv}}.
\end{equation}

The investment cost can be further divided into three parts, namely,

\begin{equation}
		C_{\mathrm{inv}}=\gamma _{\mathrm{ess}}c_\mathrm{ess}\sum_{i=1}^{N_\mathrm{ess}}S_{\mathrm{ess},i}+\gamma _{\mathrm{w}}c_\mathrm{w}\sum_{i=1}^{N_\mathrm{w}}P_{\mathrm{w},i}^\mathrm{rated}+\gamma _{\mathrm{pv}}c_\mathrm{pv}\sum_{i=1}^{N_\mathrm{pv}}P_{\mathrm{pv},i}^\mathrm{rated},
\end{equation}
{\noindent with}
\begin{equation}
    \gamma_s=\frac{\tau(1+\tau)^{T_s}}{(1+\tau)^{T_s}-1} , s=\mathrm{ess,w,pv},
\end{equation}

{\noindent where $N_\mathrm{pv}$, $N_\mathrm{w}$, $N_\mathrm{ess}$ are the number of PV stations, wind farms, and ESSs, $S_\mathrm{ess}$ is the total rated capacities of all ESSs, $P_{\mathrm{w}}^\mathrm{rated}$ and $ P_{\mathrm{pv}}^\mathrm{rated}$ are the rated power of wind and PV in an ADN, respectively. $\gamma_s$ is the capital recovery factor of equipment $s$. $c_s$ is the cost of equipment $s$ per MW or MWh. $\tau$ is the discount rate, and $T_s$ is the lifetime in units of years.}

\subsection{Operation Simulation Model}

In the simulation part, a quadratic programming (QP) model is established to determine the optimal dispatch commands that minimize $C_{\mathrm{op}}$, that is,
\begin{equation}\label{low_obj}
    \min \phantom{1} C_{\mathrm{op}} = T_\mathrm{d} (C_\mathrm{gen}+C_\mathrm{dr}+C_\mathrm{cur}+C_\mathrm{mai})  
\end{equation}
\begin{equation}
	\left\{
	\begin{aligned}
		C_{\mathrm{gen}}&=\sum_{i=1}^{N_{\mathrm{g}}}{\sum_{t=1}^T{(\mathrm{a}P_{\mathrm{g},i,t}^{2} +\mathrm{b}P_{\mathrm{g},i,t}^i +\mathrm{c})}}\\
		C_\mathrm{dr\phantom{1}} &= \sum_{i=1}^{N_\mathrm{dr}}\sum_{t=1}^{T}E_t P_{\mathrm{dr},i,t} \\
		C_\mathrm{cur} & = \delta \sum_{i=1}^{N_\mathrm{pv/w}}\sum_{t=1}^{T}(P^{\max}_{\mathrm{pv/w},i,t} - P_{\mathrm{pv/w},i,t})\\
		C_\mathrm{mai} &= \sigma \sum_{i=1}^{N_\mathrm{ess}}\sum_{t=1}^{T} \lvert P_{\mathrm{ess},i,t} \rvert
	\end{aligned}
	\right.
\end{equation}
where $N_\mathrm{g}$, $N_\mathrm{dr}$ are the number of diesel generators and demand response resources, $T_\mathrm{d}$ and $T$ are the total number of dispatch cycles in one year and the number of dispatch intervals in one dispatch cycle, which, in this case, are 365 and 24 respectively. The subscript $t$ indexes time, representing hours in a day. $P_{\mathrm{g},i,t}$ are output active power of the $i_\mathrm{th}$ diesel generator at time $t$, $\mathrm{a,b,c}$ are the generation cost index, $E_t$ is the demand cost of demand response resources at time $t$, $P_{\mathrm{pv},i,t}$ and $P_{\mathrm{w},i,t}$ are the actual consumed power from the $i_\mathrm{th}$ PV station or wind farm at time $t$, with $P^{\max}_{\mathrm{pv/w},i,t}$ being the maximum power they can generate at time $t$. $ \delta $ is the punishment factor for the curtailment of RES power, $P_{\mathrm{ess},i,t}$ is the output or input power of $i_\mathrm{th}$ ESS at time $t$, depending on its sign, and $\sigma$ is the maintenance cost factor of an ESS.

The constraints contained in the QP model are listed as follows:

\textit{1) The energy and power constraints of an ESS}

\begin{equation}
	\left\{
	\begin{aligned}
		&P_{\mathrm{ess},i,t}^{\min}\le P_{\mathrm{ess},i,t} \le P_{\mathrm{ess},i,t}^{\max}\\
		&SoC^{\min}\le SoC_t\le SoC^{\max}\\		
	\end{aligned}
	\right.
\end{equation}

\begin{equation}
	\left\{
	\begin{aligned}
		&SoC_{t+1}= (1-\eta)SoC_{t} -\lambda P_{\mathrm{ess},t}\Delta t , &P_{\mathrm{ess},t}\le 0\\
		&SoC_{t+1}= (1-\eta)SoC_{t} - \frac{P_{\mathrm{ess},t}}{\lambda}\Delta t, &P_{\mathrm{ess},t} \geq 0\\		
		&SoC_0 = SoC_T
	\end{aligned}
	\right.
\end{equation}

{\noindent where $P_{\mathrm{ess}}^{\max}, P_{\mathrm{ess}}^{\min}$ are the low boundary and up boundary of an ESS's output power and input power, respectively.  $SoC_t$ is the state of charge of an ESS at time $t$. Similarly, $SoC^{\min}, SoC^{\max}$ are the low and up boundary of the State of charge of an ESS. $\eta$ is the battery self-discharge rate, and $\lambda$ is the ESS charging/discharging efficiency.}

\textit{2) The power constraints of RESs}

\begin{equation}\label{res_cons}
	\left\{
	\begin{aligned}
		&0\le P_{\mathrm{pv},i,t}\le  P_{\mathrm{pv},i,t}^{\max} \\
		&0\le P_{\mathrm{w},i,t}\le P_{\mathrm{w},i,t}^{\max} \\
		&P_{\mathrm{pv},i,t}^{\max} = P_{\mathrm{pv},i}^{\mathrm{rated}}P_{\mathrm{pv},i,t}^* \\
		&P_{\mathrm{w},i,t}^{\max} = P_{\mathrm{w},i}^{\mathrm{rated}}P_{\mathrm{w},i,t}^* \\
	\end{aligned}
	\right.
\end{equation}

{\noindent where $P_{\mathrm{pv},i,t}^*, P_{\mathrm{wind},i,t}^*$ are the per unit values of maximum PV or wind power, dependent on the weather condition. In the actual model process, these values is randomly chosen from historical data or generated by some RES scenario simulation method. Moreover, the inequalities in \eqref{res_cons} imply that the curtailment of RES power is permitted along with a punishment cost.}

\textit{3) The linearized power flow constraints}

Generally, the typical DC power flow can hardly be used in the distribution network due to its high R/X ratio. Thus, the approximation method proposed in \cite{yuan2016novel} is adopted to linearize the power flow formulation with an acceptable error range. The simplified model is shown in \eqref{pf}.

\begin{equation}\label{pf}
	\left\{
	\begin{aligned}
		&P_{ij,t}=\frac{r_{ij}(V_{i,t} - V_{j,t}) + x_{ij} (\delta_{i,t} - \delta_{j,t})}{r_{ij}^{2}+x_{ij}^{2}}\\
		&Q_{ij,t}=\frac{-r_{ij} (\delta_{i,t} - \delta_{j,t}) + x_{ij}(V_{i,t} - V_{j,t})}{r_{ij}^{2}+x_{ij}^{2}}\\
	\end{aligned}
	\right.
\end{equation}

{\noindent where $P_{ij,t},Q_{ij,t}$ are the active and reactive power flows from bus $i$ to bus $j$ at time $t$. $V_{i,t}$ is the voltage of bus $i$ at time $t$, and $r_{ij}, x_{ij}$ are the resistance and impedance of branch $ij$, respectively.}

\textit{4) Branch power constraints }

In contrast to the traditional distribution network with a unidirectional power flow, the branch power of an ADN can be bi-directional due to the presence of massive DERs. As a result, the low boundary of branch power is less than zero in \eqref{branchpower}.

\begin{equation}\label{branchpower}
	\left\{
	\begin{aligned}
		&-P_{ij}^{\max} \le P_{ij,t} \le P_{ij}^{\max}\\
		&-Q_{ij}^{\max} \le Q_{ij,t} \le Q_{ij}^{\max}\\
	\end{aligned}
	\right.
\end{equation}

{\noindent where $ P_{ij}^{\max}, Q_{ij}^{\max}$ are the maximum allowed active and reactive power flow of branch $ij$ respectively.}

\textit{5) Power balance constraints} 

\begin{equation}\label{b_pf}
	\left\{
	\begin{aligned}
		&P_{i,t}+P_{i,t}^{\mathrm{in}} -P_{\mathrm{l},i,t} =0\\
		&Q_{i,t}+Q_{i,t}^{\mathrm{in}} -Q_{\mathrm{l},i,t} =0\\
		&P_{i,t}^{\mathrm{in}} =\sum_{i,j=1,i\ne j}^{N_{\mathrm{br}}}{P_{ij,t}}\\
		&Q_{i,t}^{\mathrm{in}} =\sum_{i,j=1,i\ne j}^{N_{\mathrm{br}}}{Q_{ij,t}}\\
	\end{aligned}
	\right.
\end{equation}

{\noindent where $P_{i,t}$, $Q_{i,t}$ are the active and reactive power injected from an ESS, RES, demand response resource, or diesel generator into bus $i$ dependent on the equipment that the bus is connected with. $P_i^{\mathrm{in}}, Q_i^{\mathrm{in}}$ are the active and reactive power injected from other buses to bus $i$ at time $t$. $P_{\mathrm{l},i,t}$,$Q_{\mathrm{l},i,t}$ are the load active and reactive power in bus $i$ at time $t$. $N_\mathrm{br}$ is the number of branches.}

\textit{6) Bus voltage constraints} 

\begin{equation}\label{node_V}
	V_{i}^{\min}\le V_{i,t}\le V_{i}^{\max}
\end{equation}

{\noindent where $V_{i,t}$ is the voltage of bus $i$ at time $t$.  $V_{i}^{\max}, V_{i}^{\min}$ are the up and low boundary of the voltage of bus $i$, respectively.}

\textit{7) Diesel generator constraints}

The diesel generators in ADNs have smaller capacities than those in the transmission network but are more flexible. Therefore, it is safe to omit the unit shutdown or opening and power climbing constraints. The capacity limit of generators is represented by a piece-wise linear equation, shown in \eqref{Slimit}, which is originally in quadratic form.

\begin{equation}
	\left\{
	\begin{aligned}
		P_{\mathrm{g},i}^{\min} \le &P_{\mathrm{g},i,t}\le P_{\mathrm{g},i,t}^{\max} \\
		Q_{\mathrm{g},i}^{\min} \le &Q_{\mathrm{g},i,t}\le Q_{\mathrm{g},i,t}^{\max} \\
	\end{aligned}
	\right.
\end{equation}
\begin{equation}\label{Slimit}
	p_k P_{\mathrm{g},i} + q_k Q_{\mathrm{g},i} <= r_k S_{\mathrm{g},i}, \phantom{1} k =1 , ... ,8		
\end{equation}

{\noindent where $	P_{\mathrm{g},i}^{\min}, Q_{\mathrm{g},i}^{\min}, P_{\mathrm{g},i}^{\max}, Q_{\mathrm{g},i}^{\max}$ are the up and low boundary of $i_\mathrm{th}$ diesel generator's output reactive and active power, and $p_k$, $q_k$, $r_k$ are the factors for the piece-wise linearization.}

\subsection{Modeling of the ADN Simulation Deviation}\label{ref_uncertainty}

For capacity planning models involving RESs, planners are inevitably confronted with the challenge of dealing with the deviation in ADN simulation caused by RES variability. To be specified, the output power of RESs in the operation simulation model is naturally a stochastic variable because of the inherent uncertainty of wind speed and solar irradiance. 
It is undeniable that accurately determining the total cost, $C_\mathrm{T}$, of an ADN that incorporates uncontrollable and stochastic sources like RESs is a challenging task.
This is because no planning approach is available to simulate all possible scenarios an ADN may encounter and construct a model entirely identical to the actual physical model.

Commonly, $C_\mathrm{T}$ is approximated by selecting a particular set of typical RES scenarios or extending the operation simulation cycle to consider as many scenarios as possible. Unfortunately, both methods fail to simultaneously pursue the algorithm's efficiency and feasibility of the final solution. For the approach proposed in the study, the simulation deviation's impact on the annual cost $C_\mathrm{T}$ of an ADN is modeled as a noise function $\epsilon(P_\mathrm{res})$, which is assumed to be in a Gaussian distribution with zero mean and $\sigma^2_n$ variance as shown in \eqref{noise}. Although the simulation deviation is unlikely to follow a perfect Gaussian distribution, the accuracy of such approximation is testified to have enough accuracy in Section \ref{Accuracy}. 

\begin{equation}\label{noisy_obj}
	C^\mathrm{ob}_\mathrm{T}(\bm x \mid  P_{\mathrm{res},i}) = C_\mathrm{T}(\bm x) + \epsilon(P_{\mathrm{res},i}) 		
\end{equation}
\begin{equation}\label{noise}
	\epsilon(P_\mathrm{res}) \sim \mathcal{N}(0,\sigma_n) 
\end{equation}

{\noindent where $C_\mathrm{T}(\bm x)$ is the actual value of the objective function corresponding to a specific capacity allocation scheme $\bm x = \{[S_{\mathrm{ess},i}]_i^{N_\mathrm{ess}}, [P_{\mathrm{w},i}^{\mathrm{rated}}]_i^{N_\mathrm{w}}, [P_{\mathrm{pv},i}^{\mathrm{rated}}]_i^{N_\mathrm{pv}} \}$. As mentioned above, it is incalculable and can only be approximated because of the simulation deviation contained in the model. $C^\mathrm{ob}_\mathrm{T}$ is the calculable value of $C_\mathrm{T}(\bm x)$ with a random RES scenario $P_{\mathrm{res},i}$ including the values of $P_{\mathrm{pv},i}^*$ and $P_{\mathrm{wind},i}^*$ in \eqref{res_cons} selected in the operation simulation. Equation \eqref{noisy_obj} indicates that the consideration of merely one scenario leads to the corruption of the simulation outcome $C_\mathrm{T}(\bm x)$ by a Gaussian noise. For practical use, the RES scenario $P_{\mathrm{res},i}$ can be randomly chosen from the historical database of RES power. If the historical data is unavailable or inadequate, the mixture density network is employed to model the volatile characteristics of RES power with the help of a neural network combined with multivariate Gaussian distributions.


Even if the observable outcome $C^\mathrm{ob}_\mathrm{T}$ is corrupted by noise, it still contains information related to its actual value $C_\mathrm{T}$. As a result, with a proper approximation of the standard deviation of noise  $\sigma_n$ and an adequate amount of outcomes, it is possible to establish a probabilistic inference model of the objective function $C_\mathrm{T}(\bm x)$ with enough accuracy. The inference model naturally turns into a powerful tool for finding the optimal capacity allocation scheme of an ADN. In the following section, the mathematical details of the inference and the entire solution process of the proposed capacity planning approach are explained.

\section{Solution Method}\label{S4}

Since all the constraints are formulated into linear ones, the remaining major handicap is to exclude the simulation deviation caused by the RES variability to reveal the actual annual cost function $C_\mathrm{T}(\bm x)$ from the obtainable noisy outcomes $C^\mathrm{ob}_\mathrm{T}(\bm x)$, as depicted in Fig. \ref{noise_influence}. 

\begin{figure}[!htbp]
	\centering
	\includegraphics[scale=1]{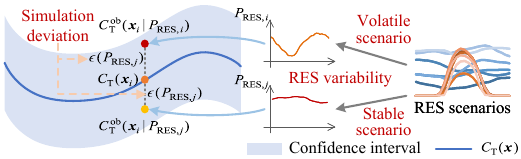}
	\caption{The influence rendered by the simulation deviation caused by the RES variability on the annual cost.}
	\label{noise_influence}
\end{figure}

To settle the problem, a modified BO algorithm, the NBO algorithm, is proposed in the study based on the methodology from \cite{letham2019constrained}. The NBO algorithm is distinct from traditional heuristic algorithms in that it generates credible inferences about the objective function through acquired outcomes and is excellently adaptive to noisy circumstances. Moreover, the NBO formulates an efficient approach to locate the latent optimal solution by its acquisition function, the Noisy Expected Improvement (NEI) function. The complete optimization process is introduced as follows:

In the beginning, a set of $N$ randomly chosen capacity allocation schemes $\bm{X_0} $ is evaluated by the operation simulation model to form an initial noisy outcome set $\bm D_0$ with their observed objective function values $ \bm y_0$, which are corrupted by Gaussian noise as explained before. And their specific meanings are shown in \eqref{inital}-\eqref{inital2}.
\begin{equation}\label{inital}
    \bm X^0 = \{\bm x_i\}_1^N = \{ [S_{\mathrm{ess},i}]_i^{N_\mathrm{ess}}, [P_{\mathrm{w},i}^{\mathrm{rated}}]_i^{N_\mathrm{w}}, [P_{\mathrm{pv}i}^{\mathrm{rated}}]_i^{N_\mathrm{pv}} \}_1^N
\end{equation}
\begin{equation}
    \bm y^0 = \{ f_i\}_1^N = \{C^\mathrm{ob}_\mathrm{T}(\bm x_i)\}^N_1 
\end{equation}
\begin{equation}\label{inital2}
	\bm D^0 = \{\bm X^0 ,\bm y^0\} 
\end{equation}

After collecting the noisy outcomes, the GP model with a fixed noise level is utilized to generate the inference of the objective function, which is referred to as the posterior distribution in the form of a multivariant Gaussian distribution, formulated by the posterior mean vector $\mu _{\bm D}$ and covariance matrix $\Sigma_{\bm D}$ shown in \eqref{GP2}, assuming the up-to-date outcome set is $\bm D=\{\bm X, \bm y\}$.

\begin{equation}\label{GP1}
    f(\bm x) \sim p( f \mid \bm D, \sigma_\mathrm{n}) = \mathcal{N}( \mu _{\bm D}, \bm \Sigma_{\bm D}) \\
\end{equation}
\begin{equation}\label{GP2}
	\left\{
	\begin{aligned}
		& \mu _{\bm D}(\bm x) = \mathrm{C} + \bm K(\bm x, \bm X)[\bm K(\bm X, \bm X) + \sigma_\mathrm{n} \mathbf I]^{-1}\bm (y-\mathrm{C}) \\
		&\bm \Sigma_{\bm D}(\bm x,\bm x^{'}) =\bm K(\bm x, \bm x^{'}) \\
		&-\bm K(\bm x,\bm X)[\bm K(\bm X, \bm X) + \sigma_\mathrm{n} \mathbf I]^{-1} +\bm K(\bm X,\bm x^{'}) \\
		&\bm K(\bm X, \bm  X)=[k(\bm x_i, \bm x_j)]_{i,j=1}^n \\
            & = [\sigma^2 \frac{2^{1-v}}{\Gamma(v)}(\frac{\sqrt{2 v}\left\|\bm x_i-\bm x_j\right\|}{\ell})^v K_v(\frac{\sqrt{2 v}\left\|\bm x_i-\bm x_j\right\|}{\ell})]_{i,j=1}^n \\
		\end{aligned}
	\right.
\end{equation}

{\noindent where $\mathbf{I}$ is the identity matrix. C is a constant. $k(\bm x_i, \bm x_j)$ is the Matérn kernel function with three kernel parameters $\sigma$, $v$ and $ l$, all determined by maximum likelihood estimation.  $\Gamma$ is the gamma function, and $K_v$ is the modified Bessel function of the second kind.}

Once the posterior distribution is established, the acquisition function, NEI, is optimized to find the latent optimal capacity allocation schemes $\hat{\bm x}$ as the candidate to be evaluated.

\begin{equation}\label{ac}
    \alpha_\mathrm{NEI}(\bm x \mid \bm D) = \int_{f^n} \alpha_\mathrm{EI}(\bm x\mid f^n)p(f^n\mid \bm D) \mathrm{d}f^n 
\end{equation}
\begin{equation}\label{EI}
	\left\{
	\begin{aligned}
		&\alpha_\mathrm{EI}(\bm x \mid \bm D) = \sigma_{\bm D}(\bm{x}) z \Phi(z) + \sigma_{\bm D}(\bm{x}) \phi(z)\\
		&z = \frac{f^* - \mu_{\bm D}(\bm{x})}{\sigma_{\bm D}(\bm{x})},\phantom{1} f^* = \min_i f(\bm x_i)
	\end{aligned}
	\right.    
\end{equation}

Although the function in \eqref{ac} has no analytic expression, it can be approximated by quasi-Monte Carol integration. The mathematical detail of the approximation in \eqref{nei} and maximization in \eqref{max_nei} can be found in \cite{letham2019constrained}.

\begin{equation}\label{nei}
     \alpha_\mathrm{NEI}(\bm x \mid \bm D) \approx \frac{1}{N}\sum^N_{k=1}\alpha_\mathrm{EI}(\bm x \mid \tilde{\bm f_k})
\end{equation}
\begin{equation}\label{max_nei}
		\hat{\bm x}= \arg \, \max \, \alpha_\mathrm{NEI}
\end{equation}

{\noindent where  $\tilde{\bm f_k}$ is a random sample from the posterior distribution $p( f \mid \bm D, \sigma_\mathrm{n})$.}

The operation simulation model evaluates the selected candidate $\hat{\bm x}$ from the optimization in \eqref{max_nei}, and the outcome set can be updated with $\hat{\bm x}$ and its corresponding annual cost $f(\hat{\bm x})$.

\begin{equation}
	\bm D^1 = \bm D^0 \cup \{\hat{\bm x}, f(\hat{\bm x})\}
\end{equation}

{\noindent whereby the updated set $\bm D_1$, a more accurate posterior inference can be generated by GP model because of the extra information about the objective function.}

Ultimately, through the iterative process of observing and inferring, it is highly guaranteed that the optimal solution can be approached. The final step is the selection of the optimum. Under the noiseless circumstances, this step is trivial since the only thing left to be done is the comparison of the values of the objective function. However, the noise brought by the simulation deviation made the decision process much more challenging. In the study, the \textit{simple reward} strategy is applied and validated, which compares the GP mean $\mu _{\bm D}(\bm x)$ in \eqref{GP2} of every evaluated capacity allocation scheme $\bm x$ to determine the optimal one.

The complete solution optimization process is illustrated in Algorithm \ref{alg:alg2}. 
\begin{algorithm}[H]
	\caption{NBO capacity planning algorithm}
	\label{alg:alg2}
	\small
	\begin{algorithmic}[1]
		\Require $\sigma_{n}$: initial value of noise standard deviation, $J$: maximum number of iterations, $\zeta$: adaptive rate.
		\Ensure $\bm x^*$: optimal capacity allocation scheme, $C_\mathrm{T}^{\min}$: optimal annual cost. 
        \Statex \textbf{Def} Sub-problem: Annual cost evaluation $f(\bm x)$
        \Statex \hspace{0.25cm} $k \leftarrow \text{random()}$ \Comment{select a random RES scenario}
        \Statex \hspace{0.25cm} $ P_{\mathrm{res}} \leftarrow \{P^*_{\mathrm{w},k}, P^*_{\mathrm{PV},k}\}$
        \Statex \hspace{0.25cm} $y \leftarrow C_\mathrm{T}^\mathrm{ob}(\bm x \mid P_{\mathrm{res}})$ \Comment{solve the QP model}
        \Statex \hspace{0.25cm} \Return $y$
        \Statex \textbf{Function} NBO($\sigma_{n}$, $J$, $\zeta$)
        \While{$i<N$}\Comment{initialize}
        \State $\bm x_i \leftarrow \text{random()}$ \Comment{generate a random scheme} 
        \State $ y_i \leftarrow f(\bm x_i)$
        \EndWhile
        \State $\bm X^0 \leftarrow \{\bm x_i\}_1^N$, $\bm y^0 \leftarrow \{ f_i\}_1^N$
        \State $j \leftarrow 0$, $\bm D^0 \leftarrow \{\bm X^0, \bm y^0\}$ 
        \State $\sigma_n^0 \leftarrow \zeta \sigma_{n} + (1-\zeta)\sqrt{\sum_{i=0}^{N+j+1}(f_i-\overline{f})^2/(N+j)} $
        \State $ p( f \mid \bm D^0, \sigma_\mathrm{n}^0) \leftarrow \mathcal{N}( \mu _{\bm D^0}, \bm \Sigma_{\bm D^0})$
	\While{$j < J$}
        \State $ \hat{\bm x} \leftarrow \max \text{ } \alpha_\mathrm{NEI}(\bm x \mid \bm D^0) $
        \State $ \hat{\bm x}\leftarrow f(\hat{\bm x}) $
        \State $ \bm D^{j+1} \leftarrow \bm D^{j} \cup \{ \bm \hat x , f(\bm \hat x)\} $ \Comment{update}
        \State $\sigma_n^{j+1} \leftarrow \zeta \sigma_{n}^{j} + (1-\zeta)\sqrt{\sum_{i=0}^{N+j+2}(f_i-\overline{f})^2/(N+j+1)} $
        \State $ p( f \mid \bm D^{j+1}, \sigma_\mathrm{n}^{j+1}) \leftarrow \mathcal{N}( \mu _{\bm D^{j+1}}, \bm \Sigma_{\bm D^{j+1}}) $
        \State $ j \leftarrow j+1$
        \EndWhile
        \State $C_\mathrm{T}^{\min} \leftarrow  \min \,  \mu _{\bm D^{J}}(\bm x)$ \Comment{simple reward strategy }
        \State $x^* \leftarrow \arg\, \min \,  \mu _{\bm D^{J}}(\bm x)$
        \State \Return $x^*$, $C_\mathrm{T}^{\min}$
	\end{algorithmic}
\end{algorithm}

\section{Case Study}\label{S5}
The case study is divided into three main parts. In the first part, the simulation deviation is eliminated by a set of typical RES scenarios, the same as most of the relevant literature does. The point of applying the conventional method is merely to prove the effectiveness of the proposed algorithm. In the second part, the proposed noise-aware modeling of simulation deviation introduced in section \ref{ref_uncertainty} is used to prove its superiority. At last, the capacity planning approach is utilized in a bigger distribution network with more elements to be configured to examine its scalability. The simulations are conducted on a computer with a 2.9 GHz Inter CORE i9-12900H processor and 32 GB of RAM, and the QP model is solved by Gurobi in Python.

\subsection{Simulation Settings}

 Relevant key parameters of the capacity planning model are listed in Table~\ref{parameter}. The planning model is built on the widely employed 33-bus distribution network model from MATPOWER with a dispatch cycle of a whole year, and the capacity allocation scheme includes the capacities of one wind farm, one PV station, and one ESS. 
 \begin{table}[htbp]
    \footnotesize
	\centering
	\caption{Key parameters of the capacity planning model}
    \begin{threeparttable}
	\begin{tabular*}{0.8\textwidth}{c@{\extracolsep{\fill}}lcc}	
		\toprule
		Equipment   & Parameter & Value & Unit \\ 
		\midrule
		\multirow{3}*{Wind farm} & Unit cost &  1,500,000   & \$/MW \\
		                         & Lifetime &   25   & year \\[3 pt]
		                         
		\multirow{3}*{PV station} & Unit cost &  500,000   & \$/MW \\
		                          & Lifetime&   25   & year \\[3 pt]
		                          
		\multirow{5}*{ESS}        & Type      &  \multicolumn{2}{c}{Lithium-ion battery}  \\
		                          & Unit cost &  125,000   & \$/MWh \\
		                          & Lifetime &   25   & year \\
	                              & Efficiency &  \multicolumn{2}{c}{$\lambda =95\% $} \\
		\bottomrule
	\end{tabular*}
     \begin{tablenotes}
        \footnotesize
        \item[a] The units for all the cost values in the following figures and tables are \$ by default.
      \end{tablenotes}
    \end{threeparttable}
\label{parameter}
\end{table}

\subsection{Effectiveness of the Proposed Approach}

This part compares the NBO algorithm with the commonly used heuristic method PSO and the basic BO. Meanwhile, five capacity allocation models are simulated with different decision variables to validate the advantage of the co-planning strategy. The following parameters are used for the PSO algorithm: the learning factor is 0.9. The maximum and minimum inertial are 0.5 and 0.3, respectively, and the number of particles is set to 10.

The average minimum annual cost derived from more than ten times repeated optimization processes of the three algorithms with 100 iterations in different capacity planning models are listed in Tab. \ref{different_config}. Besides, since in this part the RES power is represented by typical scenarios, the whole capacity planning model has no deviation and can be treated as a QP model to be solved by Gurobi. The results from solving the QP model directly serve as benchmarks for the optimization algorithms as listed in the last column of Tab. \ref{different_config}. The convergence curves of the three algorithms within a certain computation time in different models can be seen in Fig. \ref{Typical curve}. Additionally, the optimal capacity allocation schemes generated by the NBO of Model 1-4 are illustrated in Fig. \ref{Typical scheme}.
\begin{table}[!htbp]
	\centering
        \footnotesize
	\caption{The optimal cost of different capacity planning models derived from different algorithms}
	\begin{tabular*}{\textwidth}{@{\extracolsep{\fill}}ccccccc}	
		\toprule
		\multirow{2}*{Model} & \multirow{2}*{RES type} & \multirow{2}*{ESS} & \multicolumn{4}{c}{Optimum (*1e6)} \\ 
		&   && BO & NBO & PSO & QP  \\
		\midrule
		Model 0&WF only & -         &  1.17529  &  \textbf{1.17478} & 1.17479 &  1.17477\\
		Model 1&WF only & \checkmark   & 1.17903  &  \textbf{1.17662}  & {1.17671} & 1.17662 \\[3pt]
		-&PV only & -             &\multicolumn{4}{c}{Infeasible}  \\
		Model 2&PV only & \checkmark   & 1.58845 &  \textbf{1.58635} & 1.58662 & 1.58632 \\[3pt]
		Model 3&WF and PV & -       & 1.14328 & 1.13629 & \textbf{1.13628} &  1.13625\\
		Model 4&WF and PV & \checkmark & 1.13873 & \textbf{1.11947} & 1.12052 & 1.11894 \\		
		\bottomrule
	\end{tabular*}\label{different_config}
\end{table}
\begin{figure*}[!htbp]
\centering
\begin{subfigure}{0.48\textwidth}
    \includegraphics[width=\textwidth]{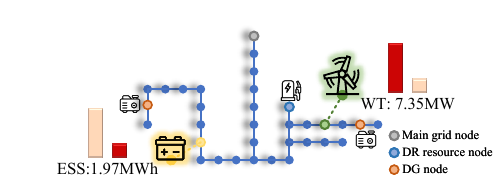}
    \caption{Model 1}
    \label{fig: dim21}
\end{subfigure}
\hfill
\begin{subfigure}{0.48\textwidth}
    \includegraphics[width=\textwidth]{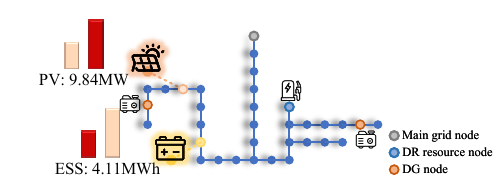}
    \caption{Model 2}
    \label{fig: dim22}
\end{subfigure}
\hfill
\begin{subfigure}{0.48\textwidth}
    \includegraphics[width=\textwidth]{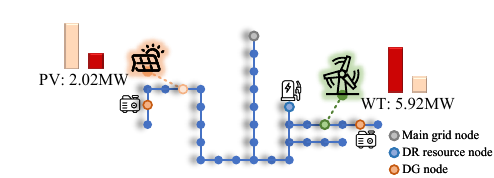}
    \caption{Model 3}
    \label{fig: dim23}
\end{subfigure}
\hfill
\begin{subfigure}{0.48\textwidth}
    \includegraphics[width=\textwidth]{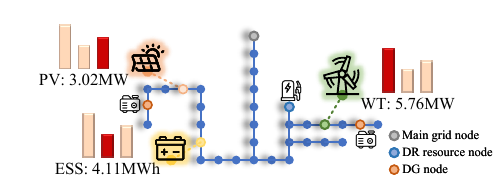}
    \caption{Model 4}
    \label{fig: dim3}
\end{subfigure}
\caption{The optimal capacity allocation scheme of Model 1-4 derived by the NBO.}
\label{Typical scheme}
\end{figure*}

\begin{figure}[!htbp]
\centering
\begin{subfigure}{0.24\textwidth}
    \includegraphics[width=\textwidth]{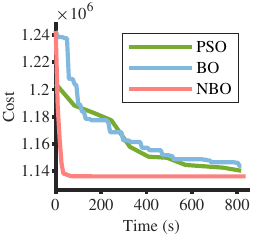}
    \caption{Model 1}
    \label{fig: resulta}
\end{subfigure}
\hfill
\begin{subfigure}{0.24\textwidth}
    \includegraphics[width=\textwidth]{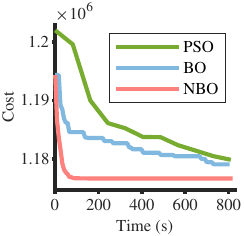}
    \caption{Model 2}
    \label{fig: resultb}
\end{subfigure}
\hfill
\begin{subfigure}{0.24\textwidth}
    \includegraphics[width=\textwidth]{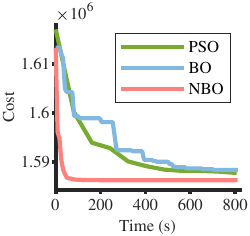}
    \caption{Model 3}
    \label{fig: resultc}
\end{subfigure}
\hfill
\begin{subfigure}{0.24\textwidth}
    \includegraphics[width=\textwidth]{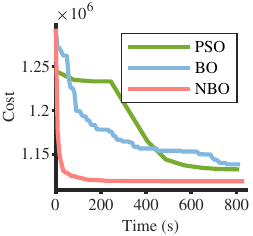}
    \caption{Model 4}
    \label{fig: resultd}
\end{subfigure}
\caption{The convergence curves of the three algorithms in different capacity planning models.}
\label{Typical curve}
\end{figure}

Based on Table \ref{different_config}, it can be observed that although the three algorithms almost converge to the same optimal values, the NBO gains a slight advantage over the other two algorithms, and its optimal values are very close to the benchmarks provided by QP. Fig. \ref{Typical curve} clearly shows that the convergence of the NBO is dramatically faster than the BO and PSO, suggesting the NBO is a more effective optimization method even under noiseless circumstances. Regarding model comparison, the collaborative capacity allocation model of a wind farm, a PV station, and an ESS achieves the lowest annual cost for an ADN. The four pictures in Fig. \ref{Typical scheme} indicate that the admission of ESSs enhances the capability of consuming RES power. Furthermore, the combination of PV and wind energy satisfies the complementary characteristics of the two RES types, reducing renewable energy curtailment and investment in return.

In order to thoroughly test the effectiveness of the model, sensitive analyses about the constraints of branch power flow and voltage in \eqref{branchpower} and \eqref{node_V} are carried out. The resulting outcome is depicted in Fig. \ref{deltapv}, and the error bars indicate the standard deviation of optimal cost from multiple repetitions of the optimization process, which have been augmented ten times for the sake of visualization. The alterations made by the parameters $\Delta P$ and $\Delta V$ to the model are formulated in \eqref{alternation1} and \eqref{alternation2}, respectively.

\begin{figure}[!htbp]
\centering
\begin{subfigure}{0.48\textwidth}
    \includegraphics[width=\textwidth]{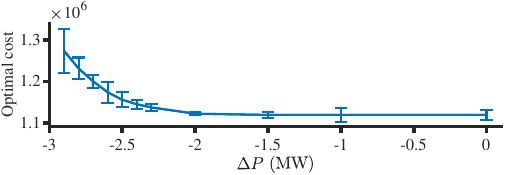}
    \caption{Case 1}
    \label{fig: deltaP}
\end{subfigure}
\hfill
\begin{subfigure}{0.48\textwidth}
    \includegraphics[width=\textwidth]{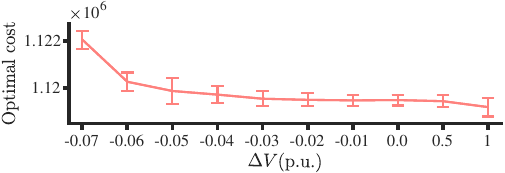}
    \caption{Case 2}
    \label{fig: deltaV}
\end{subfigure}
\caption{The trend of the average optimal cost acquired by the NBO with respect to different power flow limits and node voltage limits.}
\label{deltapv}
\end{figure}

\begin{equation}\label{alternation1}
		-(P_{ij}^{\max} + \Delta P) \le P_{ij,t} \le (P_{ij}^{\max} + \Delta P)
\end{equation}  
\begin{equation}\label{alternation2}
    	(1+\Delta V)V_{i}^{\min}\le V_{i,t}\le (1+\Delta V)V_{i}^{\max}
\end{equation}

Evidently, as the constraints become stricter, the optimal cost increases due to the need for a larger ESS capacity and increased participation of diesel generators in the operation. Both of these factors contribute to the overall rise in the annual cost.

\subsection{Superiority of the Proposed Approach}\label{Accuracy}

As claimed before, the typical scenarios are far from representing the complex operation environment of an ADN comprehensively and it is better to utilize a noise function to describe the deviation caused by RES variability. To elaborate, the typical scenario method is abandoned in the part since the essential performances of the algorithm and model have been validated. Throughout the whole planning process, each iteration of the operation simulation is carried out in a random RES scenario drawn from historical data. Aiming to testify to the adaptability and accuracy of the proposed NBO capacity planning approach along with the utility of \textit{simple reward} method under noisy circumstances, a new benchmark is set in \eqref{test_value}, referred to as test value. Unlike the benchmark from the QP model in the former subsection, no available mathematical model can be adopted to solve the problem. The test value $C^{\mathrm{test}}_{\mathrm T}(\bm x)$ has to be calculated by carrying out an examination of numerous scenarios to provide an objective indicator for the feasibility evaluation of the final capacity allocation scheme, and the scenarios that used for the examination is drawn from an independent dataset generated from the historical data with similar characteristics. The approximation error $e$ in \eqref{error} is calculated for the accuracy appraisal of the algorithms. It is worth mentioning that the calculation of the test value is extremely costly in computation and can hardly be deployed as a method to solve the problem directly. 

\begin{equation}\label{test_value}
	C_\mathrm{T}(\bm x) \approx C^{\mathrm{test}}_{\mathrm T}(\bm x) = \frac{1}{N_\mathrm{res}}\sum_{i=1}^{N_\mathrm{res}} C^{\mathrm{ob}}_{\mathrm{T}}(\bm x\mid P_{\mathrm{res},i})   
\end{equation}
\begin{equation}\label{error}
	e = \frac{\mu_{\bm D} - C^{\mathrm{test}}_{\mathrm T}}{C^{\mathrm{test}}_{\mathrm{T}}} \times 100\%
\end{equation}

{\noindent where $N_\mathrm{res}$ is set to be 200 by experiment since a bigger $N_\mathrm{res}$ will bring no significant change to the test value but only an increase in the computation time, and $\mu_{\bm D}$ is the GP mean of the objective function generated by the final GP inference model in the BO or NBO.}

As shown in Fig. \ref{sigma_comparsion}, the accuracy of the GP inference model is related to the initial noise standard deviation $\sigma_\mathrm{n}$ used in \eqref{GP2}. With the proper selection of $\sigma_\mathrm{n}$, in the case of the proposed model, optimal $\sigma_\mathrm{n}$ being 4.5e5, the error $e$ can be reduced to only -0.4251\% (BO) and 0.2044\% (NBO), acceptably for practical use. Even if when the initial value of $\sigma_\mathrm{n}$ is far from the optimal one, the error can be limited to less than 10\%, indicating great ability of the proposed approach in handling noise. Detailed information on the outcomes from BO, NBO with the optimal initial noise standard deviation ($\sigma_\mathrm{n} = 4.5\text{e}5 $) and PSO are demonstrated in Fig. \ref{noisy_pso}.

From the results in Fig. \ref{noisy_pso}, it is apparent that the NBO outperforms the BO and achieves a more stable performance since the NBO's optimum has a lower standard deviation. The vast gap between the optimal cost of PSO and the two types of Bayesian algorithms indicates that the PSO is ineffective under noisy circumstances. Such inefficiency stems from the fact that the search process of PSO directly relies on the simulation outcomes, which are easily misled by the noise.
\begin{figure}[!htbp]
\centering
\begin{subfigure}{0.48\textwidth}
    \includegraphics[width=\textwidth]{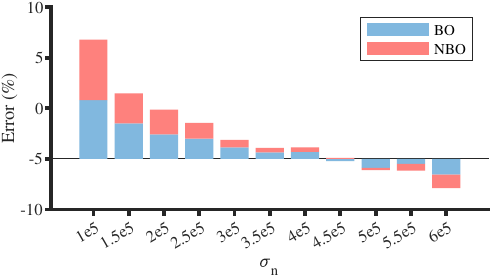}
    \caption{}
    \label{sigma_comparsion}
\end{subfigure}
\hfill
\begin{subfigure}{0.48\textwidth}
    \includegraphics[width=\textwidth]{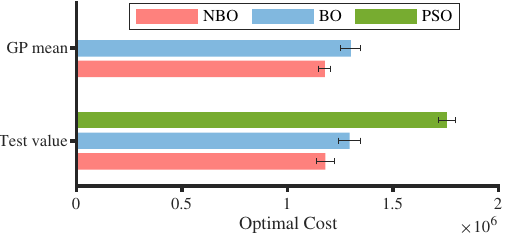}
    \caption{}
    \label{noisy_pso}
\end{subfigure}
\caption{(a) Stacked bar chart of the average error $e$ of multiple optimization processes of the NBO and BO with different values of the initial noise standard deviation $\sigma_\mathrm{n}$; (b) Bar chart of average optimal cost in both GP mean and test value of the NBO and BO  with the standard deviations of optimum acquired in each optimization shown as the error bars (the standard deviations are augmented ten times for visualization purpose).}
\end{figure}

\subsection{Scalability of the Proposed Approach}

To prove the scalability of the proposed approach, the same capacity planning process is carried out based on the 118-node distribution network model from \cite{118model}. The capacity allocation scheme included capacities for two wind farms, two PV stations, and three ESSs. Other than comparing the algorithms, the advantages of distributed and centralized configurations are also analyzed in this part, which differ in the number of nodes where all the previously mentioned elements can be installed. Specifically, centralized planning allows for installation at three nodes, whereas distributed planning enables installation at seven nodes.

Based on the curves presented in Fig. \ref{118config} and Fig. \ref{118con}, it is safe to draw the conclusion that the distributed model provides more flexibility in terms of allocation of the resources, entitling the planner to accomplish a lower annual cost and additional accommodation for RESs. On the other hand, the centralized model benefits from a lower level of optimization complexity due to a smaller number of decision variables, leading to faster convergence.
\begin{figure}[htb]
\centering
\begin{subfigure}{0.48\textwidth}
    \includegraphics[width=\textwidth]{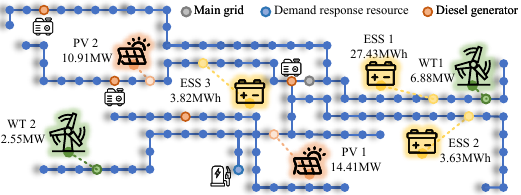}
    \caption{Distributed model}
    \label{fig: distributive}
\end{subfigure}
\hfill
\begin{subfigure}{0.48\textwidth}
    \includegraphics[width=\textwidth]{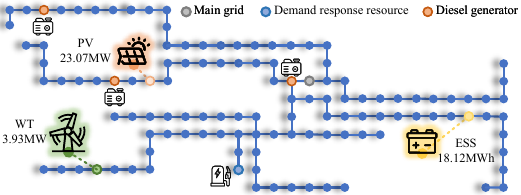}
    \caption{Centralized model}
    \label{fig: centralized}
\end{subfigure}
\caption{The optimal capacity allocation schemes of distributed and centralized models.}
\label{118config}
\end{figure}
\begin{figure}[!htbp]
	\centering
	\includegraphics[scale=1]{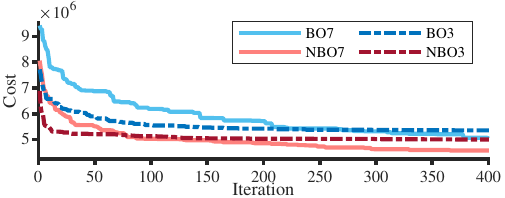}
	\caption{Convergence curves of optimal GP means for each iteration of the BO and NBO in the 118-bus model under the centralized (BO3/NBO3) and distributed planning (BO7/NBO7).}
	\label{118con}
\end{figure}

\section{Conclusion}\label{S6}
The thriving of distributed renewable generation in the distribution network requests a feasible and efficient capacity planning approach for ADNs. However, the intrinsic variability of distributed RESs in ADNs causes simulation deviation in the capacity planning model that needs to be handled.
To cope with the issue, a modeling method of the deviation is proposed in the article along with the NBO algorithm to determine the optimal capacity allocation scheme of the DERs contained in an ADN. 
The case study strongly validates that the proposed approach has a magnificent advantage in planning efficiency compared with other optimization algorithms widely used in the state-of-the-art literature, like PSO or BO. 
Even better, the NBO algorithm reveals great adaptability under noisy circumstances. Such adaptability enables the capacity planning model to effectively handle the simulation deviation engendered by the RES variability and lays a solid foundation for the feasibility of the outcome without sacrificing efficiency.
Additionally, the latent benefits of collaborative and distributed planning strategies are unrevealed in the study. The integrated complementary of two RES types and extra planning flexibility are verified to provide extra accommodation for RES power and a reduction in the annual cost of an ADN.

\bibliography{ref}

\end{document}